\documentclass[10pt,twocolumn,letterpaper]{article}

\usepackage{cvpr}              % To produce the CAMERA-READY version
\usepackage{graphicx}
\usepackage{amsmath}
\usepackage{amssymb}
\usepackage{booktabs}
\usepackage{makecell}
\usepackage{verbatim}
\usepackage{multirow}

\usepackage[pagebackref,breaklinks,colorlinks]{hyperref}

\usepackage[capitalize]{cleveref}
\crefname{section}{Sec.}{Secs.}
\Crefname{section}{Section}{Sections}
\Crefname{table}{Table}{Tables}
\crefname{table}{Tab.}{Tabs.}

\def\cvprPaperID{5016} % *** Enter the CVPR Paper ID here
\def\confName{CVPR}
\def\confYear{2022}

\begin{document}

%%%%%%%%% TITLE - PLEASE UPDATE
\title{Depth Estimation by Combining Binocular Stereo and Monocular Structured-Light}

\author{Yuhua Xu$^{1,2,}$\thanks{Corresponding author}~, Xiaoli Yang$^{2}$, Yushan Yu$^{2}$, Wei Jia$^{1}$, Zhaobi Chu$^{1}$, Yulan Guo$^{3}$
\\$^{1}$Hefei University of Technology,
$^{2}$Orbbec,
$^{3}$Sun Yat-sen University\\
 {\tt\small xyh\_nudt@163.com}
% For a paper whose authors are all at the same institution,
% omit the following lines up until the closing ``}''.
% Additional authors and addresses can be added with ``\and'',
% just like the second author.
% To save space, use either the email address or home page, not both
%\and
%Second Author\\
%Institution2\\
%First line of institution2 address\\
%{\tt\small secondauthor@i2.org}
}

\maketitle

%%%%%%%%% ABSTRACT
\begin{abstract}
It is well known that the passive stereo system cannot adapt well to weak texture objects, e.g., white walls. However, these weak texture targets are very common in indoor environments. In this paper, we present a novel stereo system, which consists of two cameras (an RGB camera and an IR camera) and an IR speckle projector. The RGB camera is used both for depth estimation and texture acquisition. The IR camera and the speckle projector can form a monocular structured-light (MSL) subsystem, while the two cameras can form a binocular stereo subsystem. The depth map generated by the MSL subsystem can provide external guidance for the stereo matching networks, which can improve the matching accuracy significantly. In order to verify the effectiveness of the proposed system, we build a prototype  and collect a test dataset in indoor scenes. The evaluation results show that the Bad 2.0 error of the proposed system is 28.2\% of the passive stereo system when the network RAFT is used.
The dataset and trained models are available at \textcolor{magenta}{https://github.com/YuhuaXu/MonoStereoFusion}.
\end{abstract}

%%%%%%%%% BODY TEXT

\section{Introduction}
Depth estimation is a fundamental problem in computer vision, which has numerous applications in the fields of 3D modeling, robotics, UAVs, augmented realities (AR), and autonomous driving~\cite{bao2020instereo2k, kitti2012, scharstein2014high}. Depth estimation methods can be divided into active structured-light, binocular stereo vision, time-of-flight (TOF), and monocular depth estimation.  

Since Microsoft Kinect \cite{zhang2012microsoft} was released in 2010, consumer-grade depth sensors have been widely used. Kinect is based on the monocular structured-light method, which was also used in iPhone X released in 2017. %So far, more than 200 million mobile phones are equipped with depth sensors. 
However, it may fail to obtain depth measurements for distant objects, or outdoor scenes under strong light. 
The binocular stereo vision system has a larger measurement range than the structured-light system, and it can also work in outdoor environment with strong sunlight, but it is easily affected by the surface texture of the objects. 
In recent years, stereo matching methods based on deep learning have achieved remarkable progress. However, these methods may still fail on scenes with weak texture (e.g., white walls). And this kind of weak texture objects are very common in indoor environment. 
The binocular active structured-light system (e.g., Intel D435 \cite{D435-web}) relies on two IR cameras and an IR projector for depth estimation, which has good adaptability in both indoor and outdoor situations. To acquire texture, a third camera (i.e. RGB camera) is required. 
Since there is a baseline between the RGB camera and IR camera, a coordinate system conversion step is required to make the depth image aligned with the RGB image. 
Due to the noise of the depth map and the error of the calibration parameters, it is difficult to accurately align the RGB image and depth map. In terms of hardware, three cameras and one projector are required, which is not compact. 
TOF has poor adaptability to objects with low reflectivity and distant objects. In addition, TOF suffers from multipath interference \cite{sarbolandi2015kinect}.
The monocular depth estimation methods cannot obtain the depth maps with a certain scale \cite{godard2019digging}.

In this work,  we seek a compact depth sensing solution that can integrate the advantages of the monocular structured-light and binocular stereo vision.

% Stereo matching is one of the key problems.  

%Recent years, deep learning methods significantly promote the progress of depth estimation methods, including stereo matching \cite{laga2020survey}, monocular depth estimation, and depth completion. 

%Please follow the steps outlined below when submitting your manuscript to the IEEE Computer Society Press.  This style guide now has several important modifications (for example, you are no longer warned against the use of sticky tape to attach your artwork to the paper), so all authors should read this new version.

The main contributions of this work are:

(1) We propose a novel stereo vision system, which consists of an RGB camera, an IR camera  and an IR speckle projector.  Especially, the IR camera is not attached with a filter. Thus the IR camera can receive IR light (invisible to human eyes) and ambient light (visible to human eyes) simultaneously. The IR camera and IR projector can form a monocular active structured-light system as Kinect, while the IR camera and the RGB camera can form a binocular stereo system. These two types of stereo systems have complementary advantages. 
The active structured-light system is robust to weak texture objects (e.g., white walls) which are hard to handle for the passive binocular stereo system. We can obtain a robust stereo system by fusing the initial depth map obtained by the active structured-light system in the cost volume of stereo matching network. 

(2) We build a prototype system and collect a new stereo dataset for integrating the monocular structured light and binocular stereo vision (MonoBinoStereo) to verify the effectiveness of the proposed method. 
The dataset will be open for further research.

(3) We find that DNN can accurately estimate the disparity map of a pair of asymmetric stereo images, where one is passive and the other is active (with speckles). 
To the best of our knowledge, this is the first time that DNN is used to process this kind of stereo images with asymmetric texture.

The features of the proposed stereo system are as follows:

(1) Compared with the classical binocular stereo vision, it is robust to weak texture objects and rich texture objects simultaneously in indoor environments.

(2) Compared with the existing monocular structured-light system (e.g., Kinect), it has a larger measuring distance range and better performance in outdoor environment.

(3) Compared with the existing active depth sensing system (e.g., Kinect and Intel D435), its output depth maps have better completeness. In addition, the depth map is naturally aligned with the RGB image pixel-by-pixel. 

(4) For the interference of strong sunlight, it will degenerate into an ordinary passive stereo system in outdoor environments.

\section{Related Work}
%\subsection{Stereo matching networks}
Zbontar \etal~\cite{zbontar2015computing} first use convolutional neural network (CNN) to compare two image patches (e.g., 9$\times$9 or 11$\times$11) and calculate their matching costs. The following steps, such as cost aggregation, disparity computation, and disparity refinement, are still traditional methods~\cite{adcensus2011building}. 
This method (i.e. MC-CNN) significantly improves the accuracy, but still struggles to produce accurate disparity results in textureless, reflective and occluded regions and is time-consuming. 
DispNetC~\cite{dispNetC2016large} is the first end-to-end stereo matching network, which is more efficient, almost 1000 times faster than MC-CNN-Acrt~\cite{zbontar2015computing}. In DispNetC, there is an explicit correlation layer. In traditional stereo matching methods, there is usually a disparity refinement module. Inspired by this, the residual refinement layers are exploited \cite{iresnet2018stereo, liang2019stereo, cascade2017} to further improve the prediction accuracy.
Besides, the segmentation information \cite{segstereo2018} and edge information \cite{edgestereo2020} are incorporated into the stereo matching networks to improve the performance. Wang \etal \cite{wang2019learning, wang2020parallax} propose a generic parallax-attention mechanism to capture stereo correspondence regardless of disparity variations. Optical flow and rectified stereo are closely related problems. 
RAFT \cite{teed2020raft} uses a gated recurrent unit (GRU) based operator to iteratively update the flow field using features retrieved from the correlation volume. RAFT shows good generalization performance.

GC-Net~\cite{gcnet2017} first uses 3D convolutions for cost aggregation in a 4D cost volume, and utilizes the soft $argmin$ to regress the disparity. 
Duggal \etal \cite{deeppruner2019} adopt the idea of PatchMatch Stereo \cite{patchmatch2011}, and build a thin cost volume to speed up the prediction process. The similar idea is also used in \cite{cascade_cost_volume2020}. Variance-based uncertainty estimation is used to adaptively adjust disparity search space of the thin cost volume \cite{cheng2020deep, shen2021cfnet}. 
%Since disparities can vary significantly for stereo cameras with different baselines, focal lengths and resolutions, the fixed maximum disparity used in cost volume hinders them to handle different stereo image pairs with large disparity variations. 
Recent work~\cite{PSMNet2018, deeppruner2019} shows that the 3D convolution can improve matching accuracy on specific datasets. 
However, 3D convolution is more time-consuming than 2D convolution, which makes it difficult to apply in real-time applications. In order to pursue real-time performance, StereoNet~\cite{stereonet2018} performs 3D convolution at a low resolution (e.g., 1/8 resolution), and then refines the disparities hierarchically. The resulting network can run in real-time at 60 fps. However, this simplification decreases the network's accuracy.
%For other networks, such as AANet~\cite{AANet2020}, FADNet~\cite{fadnet2020}, and DeepPruner-Fast~\cite{deeppruner2019}, although the accuracy has been improved, they have not achieved real-time performance (i.e. 30 fps). 

Xu \etal \cite{xu2021bilateral} design a bilateral grid based edge-preserving cost volume upsampling module. With the upsampling module, a high quality cost volume of high resolution can be obtained from the low resolution version efficiently. The upsampling module can be embedded into many existing stereo matching networks, such as GCNet \cite{gcnet2017}, PSMNet \cite{PSMNet2018} and GANet \cite{ganet2019}. The resulting networks can be accelerated by several times while maintaining comparable accuracy. HITNet \cite{tankovich2020hitnet} does not explicitly build a volume and instead relies on a fast multiresolution initialization step, differentiable 2D geometric propagation and warping mechanisms to infer disparity hypotheses. To achieve high accuracy, this method infers slanted plane hypotheses allowing to accurately perform geometric warping and upsampling operations. In order to reduce the computation burden, Yao \etal \cite{yao2021decomposition} propose a decomposition model which performs dense matching at a very low resolution (e.g., 20 $\times$ 36) and uses sparse matching at different higher resolutions to recover the disparity of lost details scale-by-scale.

ActiveStereoNet~\cite{zhang2018activestereonet} is the first deep learning solution for active stereo systems. Due to the lack of ground truth, the network is designed to be fully self-supervised. 
Instead of formulating the depth estimation via a correspondence
search problem, Riegler \etal \cite{riegler2019connecting}  show that a simple convolutional architecture is sufficient for high-quality disparity estimates in a monocular structured-light system.

%\textbf{Consumer-grade depth cameras. }
%Kinect V1 released by MicroSoft in 2010 is the first consumer-grade real-time depth camera, which has an important influence in computer vision community. Kinect V1 is based on the monocular structured-light technique \cite{zhang2012microsoft}, where the current speckle image is matched with the reference image to obtain the disparity map. The reference image is a speckle image of a planar target captured at a known depth $Z_{ref}$. The similar technique is also used in smart phones of Apple since iPhone X. 
%Kinect V2 is a depth camera based on TOF, which suffers from multipath interference \cite{sarbolandi2015kinect}. ZED \cite{ZED-web} is a popular passive stereo depth camera, which consists of two RGB cameras. 
%Intel RealSense D435 and D455 \cite{D435-web} are also binocular depth cameras. The difference is that they use two gray cameras for 3D imaging, and add a IR speckle projector to improve the reliability of 3D reconstruction in indoor scenes. The third camera, RGB camera, is used for texture acquisition.
%\subsection{Depth Completion}

Our work is also related to image guided depth completion, whose task is  to estimate the dense depth map from sparse depth measurement. 
%Uhrig \etal \cite{uhrig2017sparsity} proposed a sparsity-invariant convolution layer to enhance the depth measurements from LiDAR.
Ma \etal \cite{ma2018sparse} proposed to feed the concatenation of the sparse depth and the color image into an encoder-decoder deep network.
Jaritz \etal \cite{jaritz2018sparse} combined semantic segmentation to improve the depth completion. 
Cheng \etal \cite{cheng2018depth} proposed a convolutional spatial propagation network (CSPN) to post process the depth completion results with neighboring depth values. However, CSPN relies on fixed-local neighbors, which could be from irrelevant objects. 
Park \etal \cite{park2020non} proposed a non-local spatial propagation network for depth completion. This method can effectively avoids irrelevant local neighbors and concentrates on relevant non-local neighbors during propagation. 
Qiu \etal \cite{qiu2019deeplidar} learned surface normals as the intermediate representation. 
Xu \etal \cite{xu2019depth} modeled the geometric constraints between depth and surface normal in a diffusion module and predicted the confidence of sparse LiDAR measurements to mitigate the impact of noise.
For addressing the problem of depth smearing, Imran \etal \cite{imran2021depth} proposed a multi-hypothesis depth representation that explicitly models both foreground and background depths in the difficult occlusion-boundary regions. 
%Park \etal \cite{} .

Compared with depth completion methods, our method can utilize the stereo pair and the depth guidance from the monocular structured light subsystem for disparity estimation. When the depth guidance is not available, the stereo pair can still be used to estimate the depth of the targets. The stereo images can form stronger constraints than single images.

\section{System}

\subsection{Hardware}
\begin{figure}[h]
\centering
\includegraphics[width=0.45\textwidth]{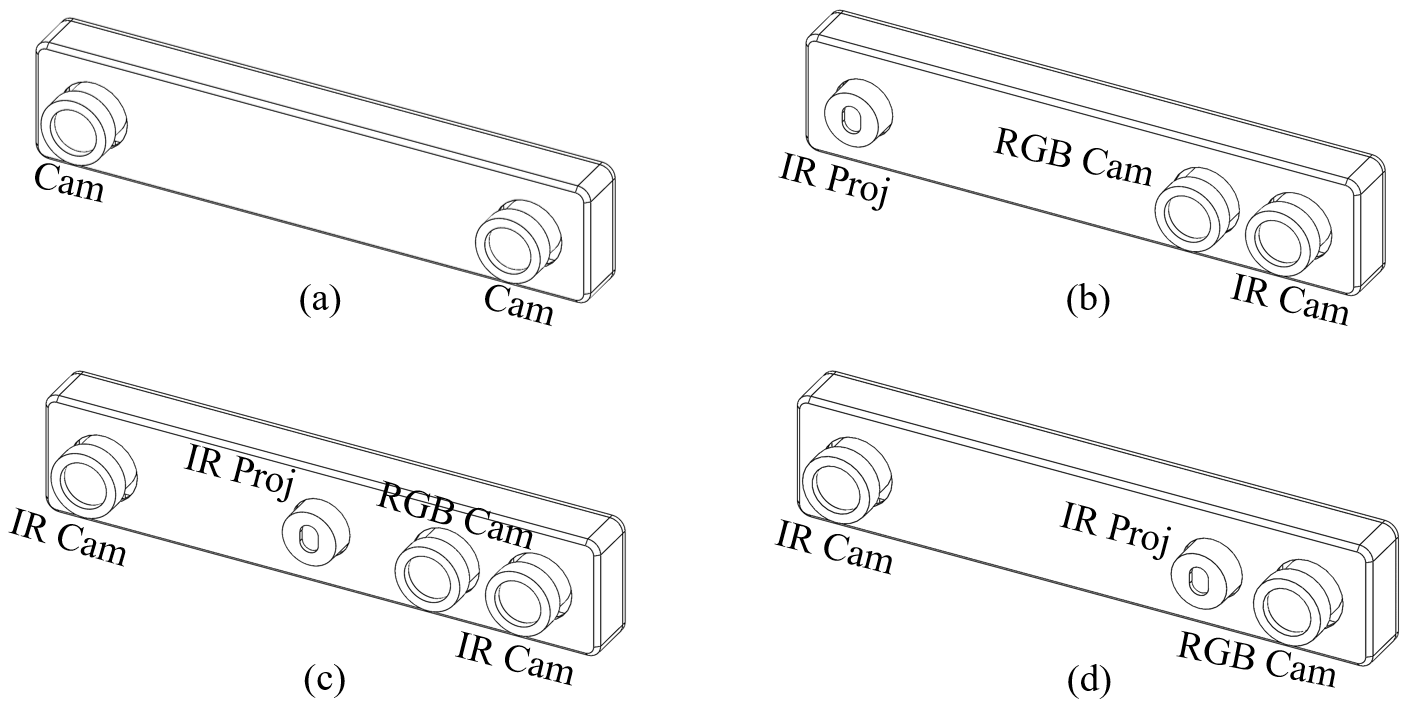}
\caption{Layout of various depth cameras. (a) Binocular stereo camera (e.g., ZED \cite{ZED-web}). (b) Monocular structured-light depth camera (e.g., Kinect \cite{zhang2012microsoft}). (c) Active binocular depth camera (e.g., Intel D435 \cite{D435-web}). (d) Design of the proposed depth camera. Compared with (b), there should be enough baseline between the IR camera and RGB camera in the proposed depth camera since the two cameras are used for the binocular stereo subsystem. In addition, the IR camera and IR projector form a monocular structured-light (MSL) subsystem. The depth map from the MSL subsystem can provide external guidance in stereo matching networks.}
\label{fig:camera_layout}
\end{figure}

In this paper, we design a novel stereo camera. As illustrated in Figure \ref{fig:camera_layout}d, the proposed stereo camera consists of an RGB camera, an IR camera and an IR projector. Its layout is similar to the monocular structured-light system (Figure \ref{fig:camera_layout}b), e.g., Kinect. However, it is significantly different from Kinect. In Kinect, the IR camera and IR projector are used for depth estimation. To obtain the depth map aligned with RGB image,  a depth-to-color step is required to convert the depth map from the IR camera coordinate system to the RGB camera coordinate system. 

The proposed stereo system consists of two subsystems. First, the IR camera and the IR projector form an active monocular structured-light subsystem. Second, the IR camera and the RGB camera form a binocular stereo subsystem. The monocular structured-light subsystem is robust to weak texture objects, while the binocular subsystem has the ability to reconstruct distant objects and can work in outdoor environment. Thus the two subsystems have complementary advantages. 

In the next subsections, we will show how the two subsystems are integrated.

\subsection{Depth Estimation Pipeline}
\begin{figure*}[htb]
\centering
\includegraphics[width=0.85\textwidth]
{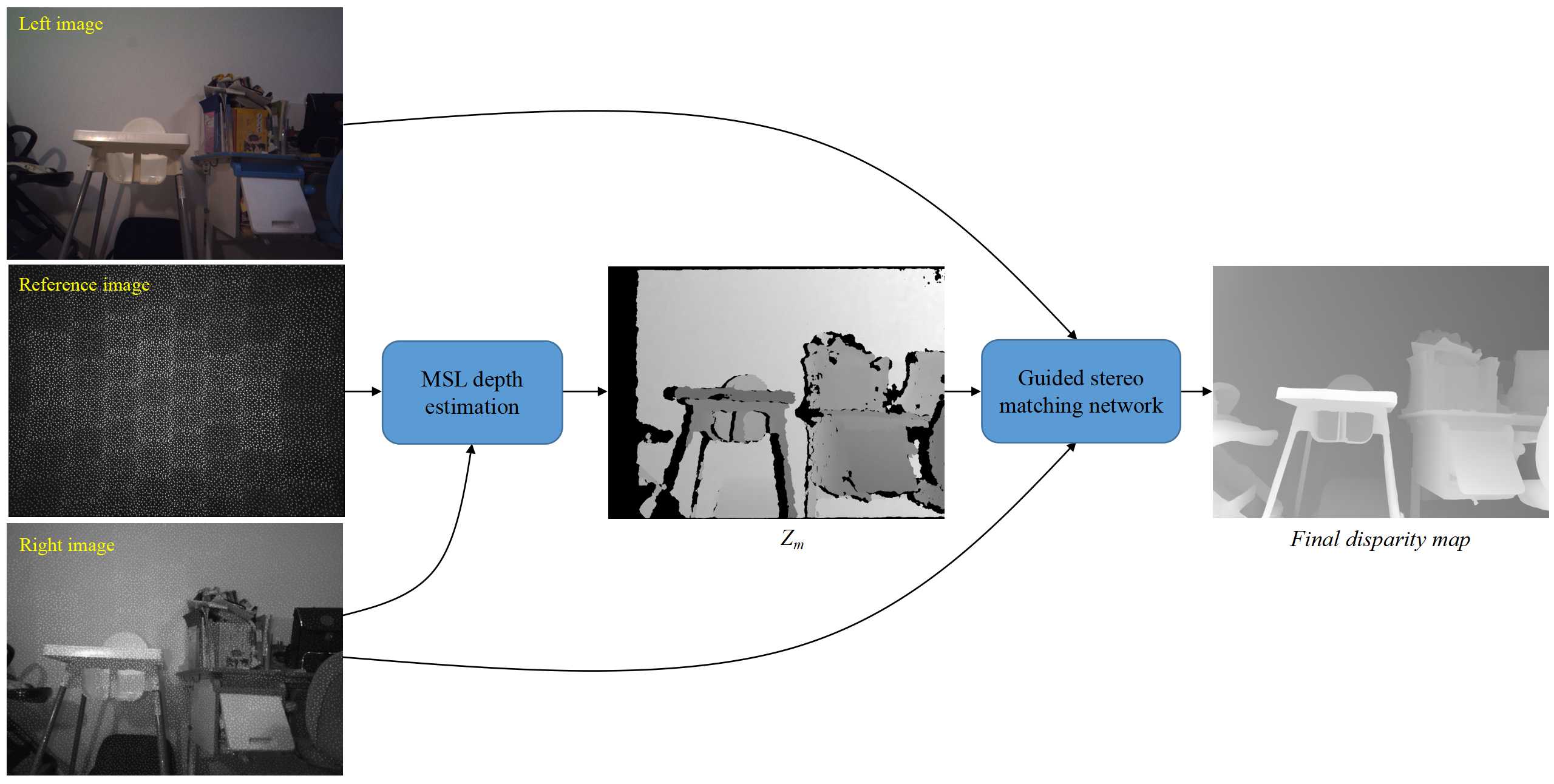}
\caption{Pipeline of the proposed depth estimation method. First, the initial depth map is obtained with the monocular structured-light (MSL) subsystem by matching the IR image and the pre-stored reference image. Then, the IR and RGB image pairs are fed to the stereo matching network to extract features and build a cost volume. The information of the active monocular subsystem is integrated in the cost volume as done in GSM \cite{poggi2019guided} to obtain high quality disparity map. }
\label{fig:pipeline}
\end{figure*}

As mentioned before, the proposed depth camera consists of two subsystems. The input includes an RGB image, an IR image and a  reference speckle image. The reference image is pre-stored and fixed in the monocular structured-light subsystem, as shown in Figure \ref{fig:pipeline}. First, the current IR image of the targets and the reference speckle image are matched, and then a disparity map $d_{m}$ is obtained. 
With the calibration parameters of the monocular structured-light subsystem, a depth map $Z_{m}$ can be obtained and re-projected to the RGB camera coordinate system. We use $Z'_{m}$ to denote the depth map aligned with the RGB image and $d'_{m}$ to denote the corresponding disparity map.
Then, the RGB image, IR image and disparity map $d'_{m}$ are fed into the stereo matching network to estimate the final disparity map. The pipeline is illustrated in Figure \ref{fig:pipeline}.

\subsection{Monocular Structured-Light}
\begin{figure}[h]
\centering
\includegraphics[width=0.35\textwidth]
{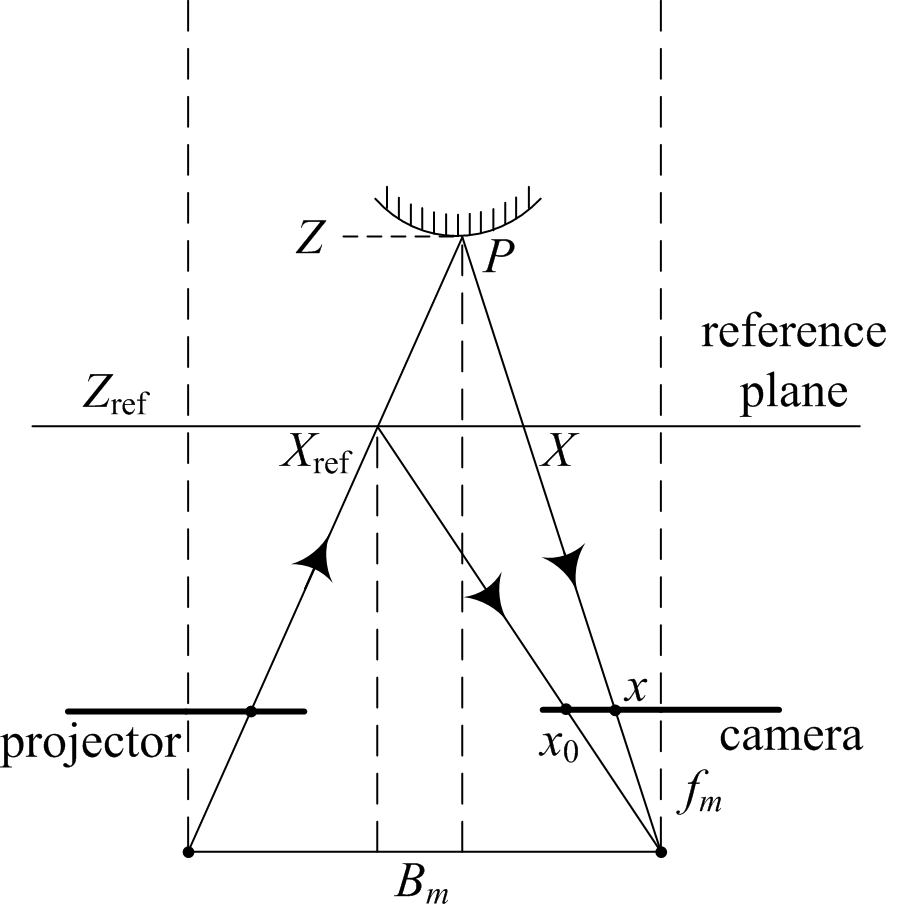}
\caption{Principle of the monocular structured-light system. The. The change in depth will bring about the movement of the speckle spots in the horizontal direction. }
\label{fig:monocular}
\end{figure}

Different from the binocular stereo system, a camera is replaced by a projector in the monocular structured-light system, as shown in Figure \ref{fig:monocular}. The depth estimation process is similar to Kinect \cite{freedman2010depth,wang2013depth}. The current speckle image of the targets is matched to the reference image, which  is a speckle image captured when the camera's optical axis is perpendicular to a planar target at a known distance $Z_{ref}$. In order to eliminate the influence of different brightness of the two images to be matched, we follow the method in \cite{wang2013depth} to convert these images to binary images. Then, an efficient block matching algorithm is used to calculate the corresponding relationships between the two images to obtain the disparity map $d_{m}$. The matching window size is set to 21$\times$21. With the disparity map, we can obtain the depth map $Z_m$ via
\begin{equation}
Z_{m}=\frac{Z_{ref}}{1-\frac{Z_{ref}d_{m} }{B_{m} f_{m}}} \\
\end{equation}
where $B_{m}$ is the baseline and $f_{m}$ is the focal length of the monocular structured-light system.
With the calibration parameters of the cameras, we can convert the depth map $Z_{m}$ onto the image plane of the RGB camera and obtain the depth map $Z'_{m}$ aligned with the RGB image. Next, we can obtain the corresponding disparity map in the binocular stereo system via 
\begin{equation}
d'_{m}={Bf}/{Z'_{m}}
\end{equation}
where $B$ is the baseline and $f$ is the focal length of the binocular system.

\subsection{Stereo Matching Network and Fusion Strategy}
Note that, the IR camera here does not have a narrow-band filter as Kinect. So the IR camera can receive the active speckled light and ambient light. Thus the images of the two cameras are very different in appearance in indoor environments, as shown in Figure \ref{fig:pipeline}. It seems that it is difficult to match this kind of images. Fortunately, we find that accurate matching results can be obtained with the deep neural networks (DNN). 

\begin{figure*}[htb]
\includegraphics[width=0.8\textwidth]{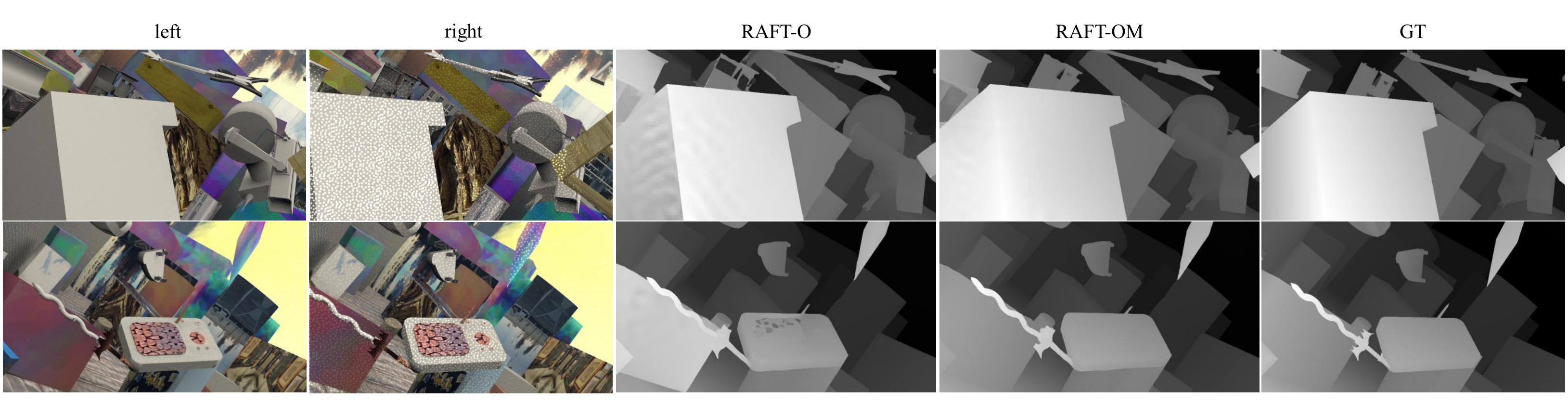}
\centering
\caption{Evaluation on SceneFlow. }
\label{fig:flying-qualitative}
\end{figure*}

In order to verify the adaptability of the DNN to this kind of binocular images with asymmetric textures, we first modify the training dataset and testing dataset of Flyingthings3D. In the modified dataset, the left image remains unchanged, while tens of thousands of random speckle dots are added in the right image, as shown in Figure \ref{fig:flying-qualitative}. So the stereo images in the modified dataset have asymmetric textures. The brightness of the speckles is decreased according to the distance from these points to the camera to mimicking energy attenuation of light energy. 
Then, we use both the original and modified training datasets to train two existing stereo matching networks, including PSMNet \cite{PSMNet2018}, and RAFT \cite{teed2020raft}. 
RAFT shows good generalization in optical flow estimation task, which requires to estimate displacement both in $X$ and $Y$ directions. Here, we make a small modification to estimate only the displacement in the $X$ direction. 

Table \ref{tab:sceneflow-eva} indicates that these networks have good adaptability to this kind of stereo images with asymmetric textures (more details are in subsection \ref{sec:quant-eva}). Figure \ref{fig:flying-qualitative} shows the qualitative results.

Although there are usually many invalid values in the depth map from the active structured-light system (Figure \ref{fig:pipeline}), the depth values are relatively reliable. So the valid depth values can be used as the guidance for the stereo matching network. 
The cost volume in stereo matching network consists of features with geometric and contextual information that allows the subsequent convolution to regress the disparity probability \cite{gcnet2017, iresnet2018stereo, PSMNet2018}. 
To integrate the advantage of the monocular structured-light system, we modify the cost volume according to the disparity map $d'_{m}$ as done in guided stereo matching (GSM)~\cite{poggi2019guided}, which peaks the correlation scores or the features activation related to the hypotheses suggested by the sparse hints and dampens the remaining ones. 

Specifically, let $g$ be a matrix of size $w\times h$, conveying the externally provided disparity values, and $v$ a binary mask, specifying which elements of $g$ are valid (i.e., if $v_{xy}$ = 1). The cost volume is denoted as $\mathcal{C}\in \mathbb{R}^{w\times h \times D_{max}\times F}$, where $D_{max}$ is the max disparity and $F$ is the feature number. Given the pixel coordinate $(x, y)$ and disparity value $g(x, y)$ from external cue $g$, GSM applies Gaussian function
\begin{equation}
f_{G S M}(x, y, d)=\lambda \cdot e^{-\frac{(d-g(x, y))^{2}}{2 \sigma^{2}}}
\end{equation}
on the features $\mathcal{C}(x,y,d)$ of the cost volume, and obtain a new cost volume $\mathcal{C'}$,
\begin{equation}
\mathcal{C'}(x,y,d)=\left(1-v_{x y}+v_{x y} \cdot f_{G S M}(x, y, d)\right) \cdot \mathcal{C}(x,y,d)
\end{equation}
where $\sigma$ determines the width of the Gaussian, while $\lambda$ represents its maximum magnitude and should be greater than or equal to 1.

For RAFT, the correlation values in the cost volume are normalized to $[0,1]$ to avoid peak negative correlations via
\begin{equation}
\mathcal{C}(x,y, d)=\frac{<F_l(x,y),F_r(x-d,y)>}{2(||F_l(x,y)||+\epsilon)(||F_r(x-d,y)||+\epsilon)} + 0.5
\end{equation}
where, $F_l$ and $F_r$ are features extracted from the left and right images, $d$ denotes the disparity, and $\epsilon$ is a small constant.

In this work, the disparity map $d'_{m}$ is taken as the external guidance for the stereo matching networks.

% 拿PSMNet网络来举一个例子，展示如何修改代价空间

% 在Flyingthings的测试
\begin{table} 
\begin{center}
\begin{tabular}{|c|c|c|}
\hline
Method    & \makecell{EPE\\(Original)} & \makecell{EPE\\(Modified)} \\
\hline
PSMNet-O\cite{PSMNet2018} & 0.895	& 3.922 \\
PSMNet-M & 1.212	& 0.955\\ 
PSMNet-OM & 0.925	& 0.984 \\ 
PSMNet-OM-G & 0.666 &	0.686\\
\hline
RAFT-O\cite{teed2020raft} &  0.985	 & 1.910\\
RAFT-M &  1.070	&1.092\\ 
RAFT-OM & 1.026	& 1.109\\
RAFT-OM-G & 0.751 & 0.771\\
\hline
%ZZNet-O & 1/4 & \\
%ZZNet-OM & 1/8 & \\ 
%\hline
%SGBM [] &  & \\
\end{tabular}
\end{center}
\caption{Evaluation of networks on the original SceneFlow dataset and the modified SceneFlow dataset. We use suffixes O, M and OM to denote the models trained with the original Flyingthings3D dataset, the modified Flyingthings3D dataset and the mixture of the two datasets, respectively.  The suffix G denotes the guidance is used in the network.}
\label{tab:sceneflow-eva} 
\end{table}

% realscene测试
\begin{table*} [htb]
\begin{center}
\begin{tabular}{|c|c|c|c|c|c|c|c|c|}
\hline
%Method  & \makecell{EPE\\(DOE-On)} & \makecell{Bad0.5 (\%)\\ (DOE-On)} & \makecell{Bad1.0(\%)\\(DOE-On) } & \makecell{Bad2.0(\%)\\(DOE-On)}  
%& \makecell{EPE\\(DOE-Off)} & \makecell{Bad0.5 (\%)\\ (DOE-Off)} & \makecell{Bad1.0(\%)\\(DOE-Off) } & \makecell{Bad2.0(\%)\\(DOE-Off)}\\
\multirow{2}{*}{Method} & \multicolumn{4}{c|}{projector on}  & \multicolumn{4}{c|}{projector off}\\
        \cline{2-5}  \cline{6-9}
        & EPE & Bad0.5 (\%)  & Bad1.0 (\%) & Bad2.0 (\%) & EPE & Bad0.5 (\%)  & Bad1.0 (\%) & Bad2.0 (\%)\\
\hline
%PSMNet-OM &  & & &\\
PSMNet-O \cite{PSMNet2018} & 9.007	& 70.51 &	55.66	& 41.79& 2.112 &	52.54	& 33.85 &	20.42 \\ %无引导
PSMNet-OM & 2.687	& 57.35	& 39.28	& 24.81 &	\textbf{1.871}	& \textbf{51.70}	& 33.16 &	19.89\\
PSMNet-OM-G &  \textbf{{0.814}} & \textbf{{45.63}} &\textbf{{ 15.73}} & \textbf{{3.81}} & 2.018 &	52.02 &	\textbf{32.67}	& \textbf{18.33} \\ %有引导训练的模型
\hline
RAFT-O \cite{teed2020raft} & 2.498	& 57.83	& 37.70 &	21.88 & 1.183	& 46.43 & 26.07 & 12.71\\
RAFT-OM & 1.370	& 49.23	& 29.31	& 14.60 & 1.239	& \textbf{44.21}	& \textbf{23.18}	& 11.72\\ 
{RAFT-OM-G} & \textbf{{0.811}} & \textbf{{45.13}} & \textbf{{16.08}}	& \textbf{{3.59}} & \textbf{1.103}	& 44.75	& 23.71 & \textbf{10.51}\\
\hline
%SGBM [] &  & &\\
%\hline
%SL [] &  & &\\
MSG \cite{li2020msg} & {3.092}	& {58.85}	& {30.32} &	{14.25}  & -	& -	& -	& {-}\\

\hline
\end{tabular}
\end{center}
\caption{Quantitative evaluation on the real scene dataset. The suffix G denotes the guidance is used during training of the network models. Note that, when the projector is on, depth from MSL is used as the guidance in the models with suffix G. When the DOE projector is off (i.e., both the left and right images are passive), the guidance is not available and not used in network prediction. }
\label{tab:realscene-eva} 
\end{table*}

%-------------------------------------------------------------------------
\section{Experiments}
\subsection{Prototype}
\begin{figure}[h]
\includegraphics[width=0.35\textwidth]{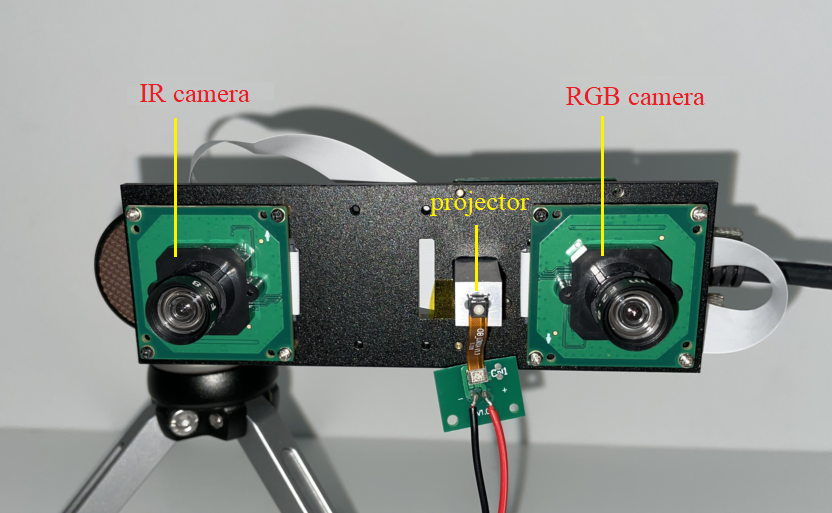}
\centering
\caption{Prototype of the proposed depth camera. }
\label{fig:prototype}
\end{figure}

To verify the effectiveness of the proposed system, we build a prototype system as shown in Figure \ref{fig:prototype}. The system includes two synchronized CMOS cameras and an IR speckle projector. Both cameras have a focal length of 4.0 mm and a resolution of 1280$\times$960. The maximum frame rate is 30 frames per second (fps). The baseline of the stereo subsystem is 94.14 mm and that of monocular structured-light system is 63.0 mm. The diffractive optical element (DOE) based projector can project about 11,000 speckle dots onto the scenes. This kind of projector is very cheap (less than \$3). We capture a speckle image of a white wall as the reference image at a distance of 80 cm when the optical axis of the camera is perpendicular to the white wall. The RGB camera has an IR-cut filter, and the IR camera has no filter. 

\subsection{Dataset and  Evaluation Metrics} \label{sec:dataset}

\textbf{Synthetic dataset. }
The synthetic SceneFlow \cite{dispNetC2016large} stereo dataset includes Flyingthings3D, Driving, and Monkaa. The dataset consists of 35,454 training images and 4,370 testing images of size $960\times 540$ with accurate ground-truth disparity maps. We will use Flyingthings3D for study of the stereo matching networks. The End-Point-Error (EPE) will be used as the evaluation metric.

\begin{figure*}[htb]
\includegraphics[width=0.78\textwidth]{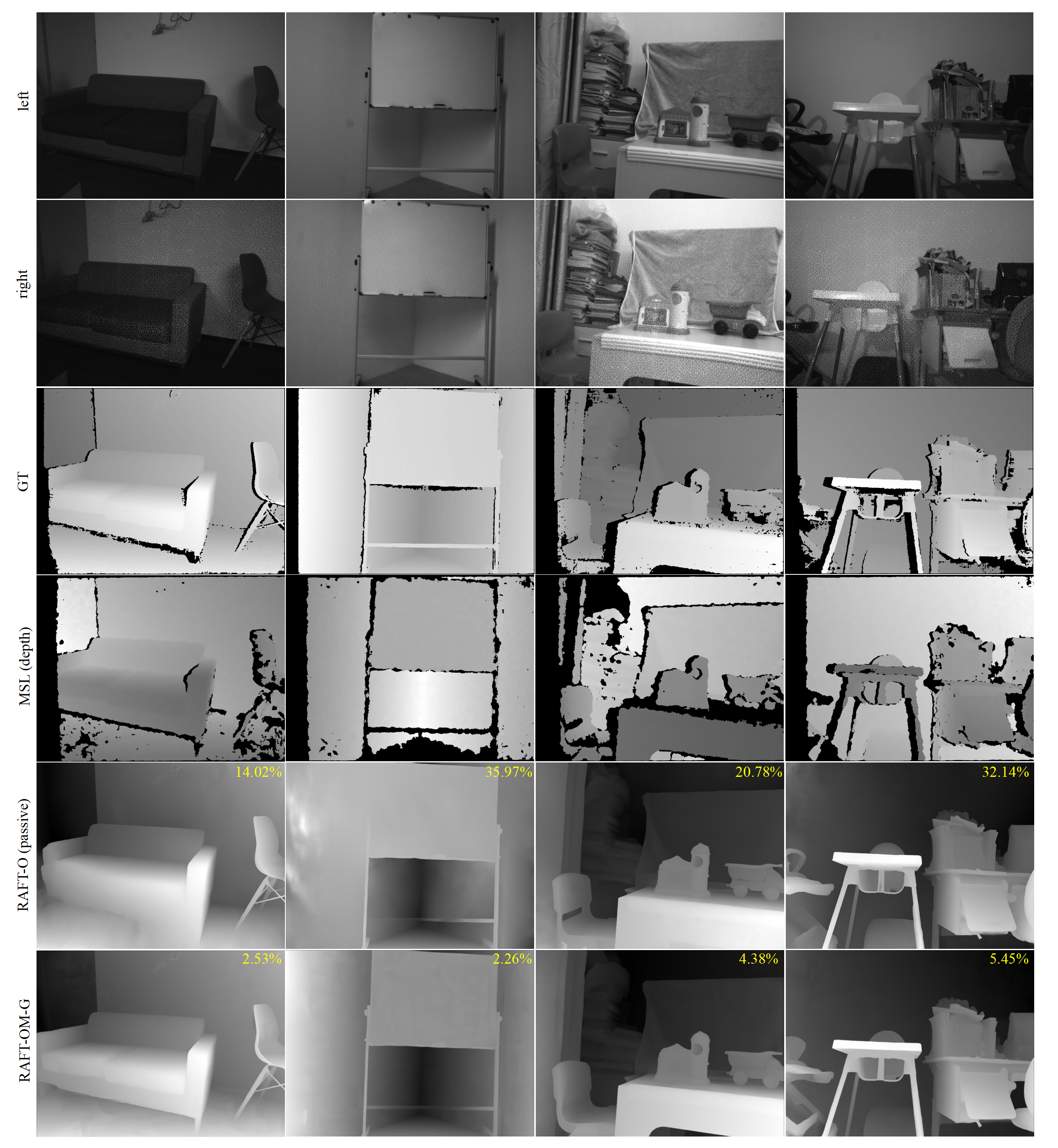}
\centering
\caption{Comparisons on the real dataset. The first row shows the left images (The RGB images are converted to grayscale images before network prediction). The second row shows the right images with speckles (the passive right images are not shown), the third row is the ground truth disparity maps generated with the space-time stereo method \cite{zhang2003spacetime,davis2003spacetime}, the fourth row shows the depth images generated with the MSL subsystem, the fifth row shows the disparity maps of RAFT-O for the passive stereo images, and the last row shows the disparity maps of RAFT-OM-G, where the left image is passive and the right image is with speckles. In row 5 and row 6, Bad2.0 error is shown for each disparity map. The corresponding error maps are shown in the supplementary material.}
\label{fig:test_dataset}
\end{figure*}

\textbf{Real-scene dataset. }
To evaluate the performance of the proposed system, we collect a dataset (i.e., MonoBinoStereo) in indoor environment, which covers different indoor scenes, including offices, living rooms, and bedrooms. The stereo pairs are easy to acquire. However, it is not a easy task to acquire the corresponding ground-truth disparity maps for the stereo pairs. 
Here, we choose to use the space-time stereo method~\cite{zhang2003spacetime, davis2003spacetime} to obtain the  ground truth disparities as done in \cite{dal2015probabilistic}. 
200 pairs of stereo images are captured for each scene. During the process of image capturing, thousands of moving speckles are projected. Therefore, the speckle distribution in each frame is different. The ground truth disparity maps are estimated by integrating all the 200 pairs of images. 
A sub-pixel refinement and a left-right check (LRC) are also applied. The MonoBinoStereo dataset includes 15 scenes in total. The samples are shown in Figure \ref{fig:test_dataset}. 
For each scene, we collect two stereo pairs, where the left images are always passive, while one image of the right camera is passive (with projector off) and the other is active (with projector on).

However, we lack a large training dataset in real indoor scenes. The synthetic IRS dataset \cite{IRS2019} is considerably close to the real scenes. It contains more than 100,000 pairs of 960$\times$540 resolution stereo images (84,946 for training and 15,079 for testing) in indoor scenes.  We use the IRS dataset as the training dataset for evaluation on the MonoBinoStereo dataset. Details of the network training are presented in the supplementary material. 

\subsection{Quantitative Evaluation}\label{sec:quant-eva}

We first evaluate the proposed method on the SceneFlow dataset. We trained PSMNet \cite{PSMNet2018} and RAFT \cite{teed2020raft} with the original Flyingthings3D dataset and the modified Flyingthings3D dataset respectively. 
%We use PSMNet-O and RAFT-O to denote the models trained with the original Flyingthings3D dataset, PSMNet-M and RAFT-M for the models trained with the modified Flyingthings3D dataset, PSMNet-OM and RAFT-OM for the models trained by mixing the two datasets. 
%The  models trained with the original datasets are denoted with a suffix O (e.g, PSMNet-O), and the models trained with the model
We use suffixes O, M and OM (e.g, PSMNet-O) to denote the models trained with the original Flyingthings3D dataset, the modified Flyingthings3D dataset and the mixture of the two datasets, respectively. The End-Point-Error (EPE) results are reported in Table \ref{tab:sceneflow-eva}. When the models are trained with the original dataset, the EPEs on the modified test dataset are large. For example, the EPE of the PSMNet-O on the modified test dataset is 3.922. When the modified training dataset is used, the EPE of the resulting model (PSMNet-M) is reduced to 0.955. However, the EPE for the original test dataset increases from 0.895 to 1.212. 
When both  training datasets are used, the resulting model (PSMNet-OM) can balance the two test datasets. 
Furthermore, if the external guidance is available, we can use the strategy in GSM \cite{poggi2019guided} to further improve the results. The resulting methods are denoted with a suffix G, e.g., PSMNet-OM-G. 
%Note that PSMNet-OM-G and PSMNet-OM represent the same model. 
%The difference is whether external guidance is used. 
When 5\% pixels of the ground truth depth map are used as the external guidance, the EPE is reduced from 0.984 to 0.686 on the modified test dataset. 
The results are similar for RAFT. The qualitative results are shown in Figure \ref{fig:flying-qualitative}. 

\begin{figure}[h]
\includegraphics[width=0.47\textwidth]{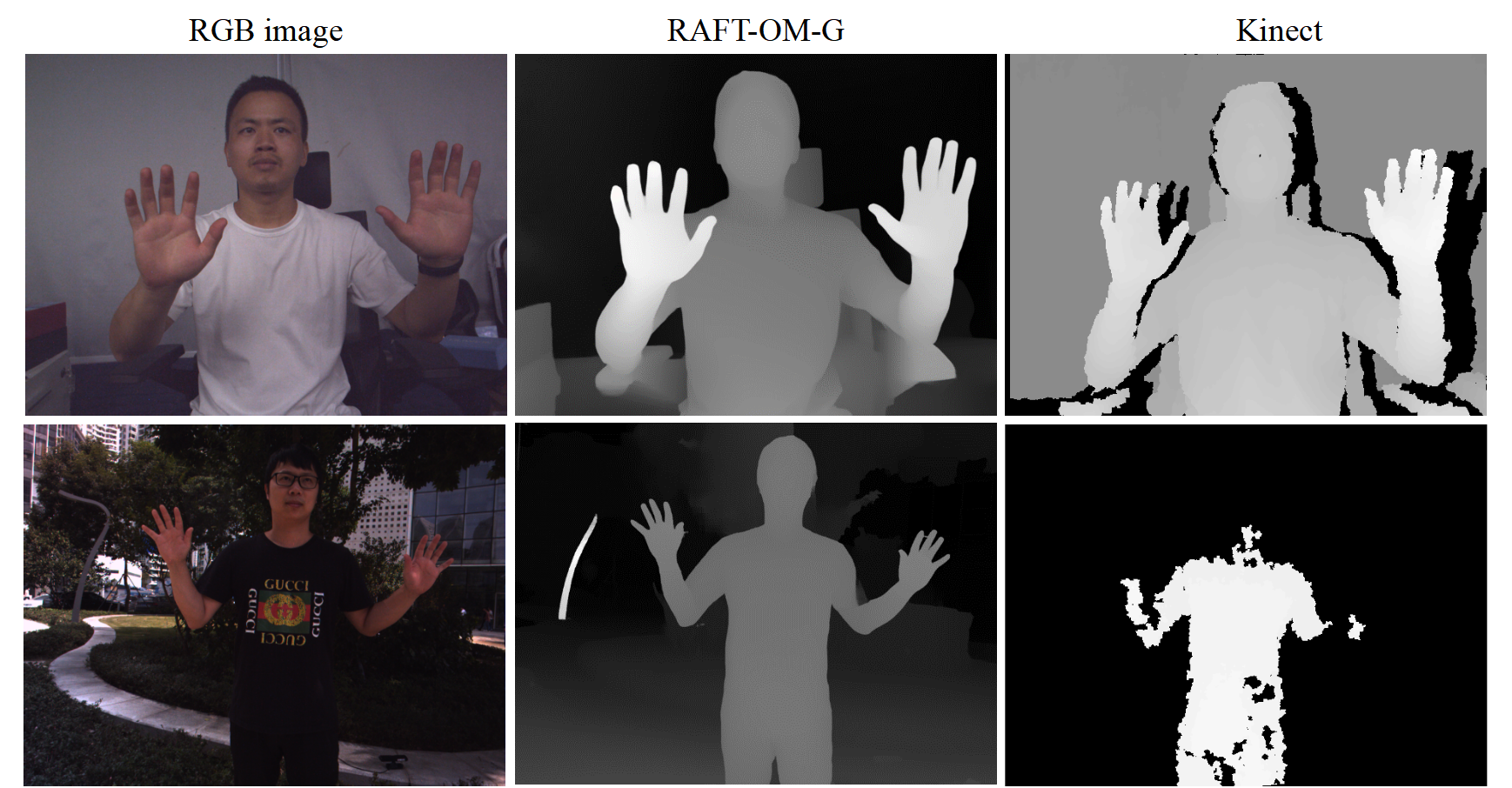}
\centering
\caption{Qualitative comparison. The first column shows the RGB images, the second column shows the disparity maps of RAFT-OM-G, the third column shows disparity maps of Kinect. The first row is the results in indoor scenes, and the second row shows the results in outdoor scenes. It is difficult for Kinect to output stable depth map out of doors. To keep anonymous, the faces are masked.}
\label{fig:Qualitative-com}
\end{figure}

\begin{figure}[h]
\includegraphics[width=0.46\textwidth]{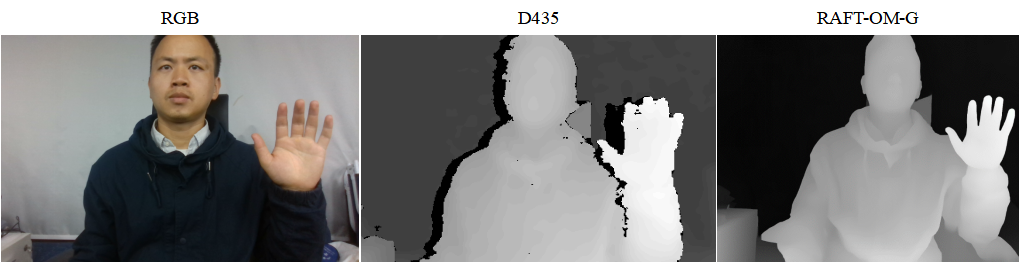}
\centering
\caption{Qualitative comparison with Intel RealSense D435 \cite{D435-web}. D435 uses two cameras to obtain depth map and a third camera for texture acquiring, where occlusion is inevitable. In contrast, Our system can output depth maps naturally aligned with RGB images with only two cameras. To keep anonymous, the face is masked.}
\label{fig:d435-com}
\end{figure}

To further verify the effectiveness of the proposed method, we evaluate the models on the collected real-scene dataset, MonoBinoStereo. The models are trained by mixing the Flyingthings3D and IRS datasets.
The quantitative results are shown in Table \ref{tab:realscene-eva}. Take RAFT for example. 
The Bad 2.0 error of RAFT-O is up to 21.88\% on the real test dataset with the DOE projector on, where only the original datasets (Flyingthings3D and IRS) are used for training. When the modified datasets are added, the Bad 2.0 error of the resulting model (RAFT-OM) is reduced to 14.60\%. 
In our system, the depth map from the monocular structured-light subsystem can be used as the external guidance for the stereo matching networks. 
We use 10\% of the pixels in $d'_m$ as the guidance\footnote{Since the cost volume is built at 1/8 resolution for RAFT, only 1/640 of the pixels in $d'_m$ are used for guidance actually.}. 
When this guidance is utilized, the Bad 2.0 error is reduced to 3.59\% (RAFT-OM-G). 
In Table \ref{tab:realscene-eva}, the quantitative results of the different models on the pure passive stereo dataset (see Subsection \ref{sec:dataset}) are also shown. 
Note that the guidance information of the passive mode is not available. 
We run the model RAFT-O on the passive test dataset. The Bad 2.0 error is 12.71\%, which is 3.5 times of RAFT-OM-G. It indicates that the proposed method can improve stereo matching accuracy significantly.
The Bad 2.0 error on the passive dataset of RAFT-OM-G is 10.51\% (no guidance used), which indicates that RAFT-OM-G can be well generalized to passive scenes. 
The qualitative results are shown in Figure \ref{fig:test_dataset}. 
Table \ref{tab:realscene-eva}  also shows that the overall performance of RAFT is better than PSMNet on MonoBinoStereo. 

In addition, we compare with a depth completion method, MSG\cite{li2020msg}, on the MonoBinoStereo dataset, where 1\% of the pixels in $d'_m$  are used as the guidance. The results are shown in Table \ref{tab:realscene-eva}. The Bad 2.0 error of MSG is 18.57\%, which is much larger than RAFT-OM-G. 

\subsection{Qualitative Evaluation}

We also test the proposed system in dynamic scenes with people and outdoor scenes, where it is difficult to obtain the ground truth disparity maps. For these scenes, we present the qualitative comparison results.
%Figure \ref{fig:Qualitative-com} shows the comparisons between the proposed fusion method and the passive stereo matching method. 
%For regions with weak texture, the estimated disparities deviates seriously for the passive stereo matching method (RAFT-O), while the foreground and background objects are well distinguished from the depth maps obtained by the proposed method. 

In Figure \ref{fig:Qualitative-com}, we compare the proposed system with Kinect V1 in indoor and outdoor scenes. 
Kinect can output dense depth estimation in indoor scenes. However, in outdoor scenes, there are more holes in the depth maps because the IR speckles projected are interfered by the sun light. 
%For the depth maps of Kinect, there are many holes for the regions of distant objects and objects with low albedo. 
However, for the proposed system, it will degenerate into a passive binocular stereo system, where the stereo pairs can still be used to estimate the dense depth maps of the scenes. 
We also compare our system with Intel RealSense D435 \cite{D435-web}, the results are shown in Figure \ref{fig:d435-com}.

\subsection{Limitation}
In the monocular structured light system, a reference image of a planar target with known depth $Z_{ref}$ is required. 
When capturing the reference image, we assume that the optical axis of the camera is perpendicular to the planar target, which is hard to guarantee in practice.
Compared with the binocular stereo system, the monocular structured light system is more difficult to calibrate. 
The calibration error will lead to alignment error of the RGB image and the depth image $Z'_m$, which may cause wrong guidance in guided stereo matching network. 
In experiment, we find that increasing the number of guide points does not improve the accuracy (see supplementary material for details).
Furthermore, if the same number of guidance points are sampled from the ground truth, the Bad0.5, Bad1.0 and Bad 2.0 errors are reduced to 12.94, 4.94, and	2.00 for RAFT-OM-G, respectively.
%There are few relevant literatures. 
So in the future, we will focus on the accurate calibration method of the monocular structured light system to further improve performance. 

\section{Conclusion}

In this paper, we present a novel stereo system. This system includes a monocular structured-light subsystem and a binocular stereo subsystem. 
These two subsystems are combined to obtain robust depth estimation. 
Our system is unique in that it has only two cameras, an RGB camera and an IR camera. The RGB camera is used both for depth estimation and texture acquisition. The depth maps obtained are naturally aligned with RGB images pixel-by-pixel. 
We collect a real test dataset in indoor scenes. The quantitative results show that the Bad 2.0 error of the proposed system is 28.2\% of the classical passive stereo system. Under strong outdoor light, the proposed system will degenerate to a passive stereo system. 
We hope the proposed system can provide a new solution for designing more robust depth cameras for the community.

\textbf{Acknowledgements.} This work was  supported by Orbbec Inc. (No. W2020JSKF0547), and partly supported by the National Natural Science Foundation of China (No. U20A20185, 61972435, 62076086), Major Science and Technology Projects in Anhui Province (202103a05020001), and Key Research and Development Program in Anhui Province (202004d07020008).

%%%%%%%%% REFERENCES
{\small
\bibliographystyle{ieee_fullname}
\bibliography{egbib}
}

\end{document}